\documentclass{article}

\usepackage[english]{babel}
\usepackage{algorithm,algorithmic}

\usepackage{graphicx}

\title{Revolutionary Algorithms}
\author{Ronald Hochreiter and Christoph Waldhauser}
\date{April 2012}

\begin{document}

\maketitle

\begin{abstract}
The optimization of dynamic problems is both widespread and difficult. When conducting dynamic optimization, a balance between reinitialization and computational expense has to be found. There are multiple approaches to this. In parallel genetic algorithms, multiple sub-populations concurrently try to optimize a potentially dynamic problem. But as the number of sub-population increases, their efficiency decreases. Cultural algorithms provide a framework that has the potential to make optimizations more efficient. But they adapt slowly to changing environments. We thus suggest a confluence of these approaches: revolutionary algorithms. These algorithms seek to extend the evolutionary and cultural aspects of the former to approaches with a notion of the political. By modeling how belief systems are changed by means of revolution, these algorithms provide a framework to model and optimize dynamic problems in an efficient fashion.
\end{abstract}


\section{Introduction}
In evolutionary computing, the treatment of dynamic problems is troublesome. Dynamic problems are characterized by their constant changing of the optimum of a target function sought to be optimized. Unfortunately, many problems in the real world are dynamic: optimal routing solutions for mail delivery as new pieces are arrive \cite{dorigo1999ant}, finding the best path through a volatile landscape \cite{kurzhanski2001dynamic}, portfolio optimization while markets are trading \cite{pliska1986stochastic}, election forecasting while new polling stations are being counted \cite{hochreiter2011evolved}. These are a few examples of dynamic problems. Clearly, if the update frequency of any such problem is low, traditional evolutionary methods can be used to find any number of successive optima, simply by restarting the optimization process on the basis of new data. However, as the update frequency increases, this approach becomes more and more impractical. Once the data is updated at an interval that is smaller than 
the time frame needed to complete the optimization, a breakdown point is reached and other ways of obtaining the successive optima need to be derived.

A possibility of postponing the advent of the breakdown point is the application of algorithms that can be parallelized \cite{cantu2000efficient}. By executing the optimization process on multiple nodes concurrently, the time required for an optimization can be reduced by a factor almost proportionally to the number of nodes. In practice, this is often realized by using multiple populations in a genetic algorithm. The drawback of these parallel algorithms is that the quality of the overall solution suffers, as multiple smaller populations also limit the gene pool for any instance of that algorithm. Thus, the risk of becoming trapped in local optima is increased.

One way to mitigate the problem of local optima is the use of cultural algorithms \cite{reynolds1994introduction}. There, knowledge about the search is stored in a belief space that is influenced by the most successful solutions and which in turn influences all children. However, the implementation of cultural algorithms in a parallel, multi-population setting is not trivial. Either there is a global belief space or separate belief spaces for each population. In the former case, execution runtime will be dictated by the slowest node, as only after all generations have finished their evaluation, the best solution can be identified to update the belief space. In the latter case, the merging of different belief spaces to contain the best solution is rather complex and application specific.

This paper deals with suggesting an entire new class of algorithms that can be used to solve dynamic problems in a parallel fashion, using multiple populations and belief spaces, without the need to merge these belief spaces. The proposed revolutionary algorithms extend the concept of cultural algorithms by evolving culture into politics. In the real world, belief systems are relatively constant over time. However, when a belief system fails to inspire its followers to achieve greater goals, when it fails to make good for what it promises, a revolution might eradicate the belief system itself and replace it with new content.

Revolutionary algorithms seek to emulate this human behavior in their goals to optimize a function. While being comparable to cultural algorithms at first glance, there are important differences. Revolutionary algorithms allow for some solutions to exist and evolve beyond the influence of the belief system. This subculture either prospers or withers and dies. If the alternative path of the subculture becomes more successful than the main line inspired by the belief system, the subculture takes over the belief system and installs its own values. Thereby the former subculture establishes hegemonic reign over all other solutions, spreading its influence. Until a new subculture evolves and eventually takes over. 

This approach promises to successfully evade local minima by always also trying out unorthodox solutions and to rapidly adapt to dynamic problems. Revolutionary algorithms thereby provide a class of algorithms that can be applied to solve problems previously intractable. 

\section{Overview of Cultural Algorithms}
Research in genetic algorithms has produced a vast number of subtypes \cite{haupt1998practical}. In the following, the focus will be put on cultural algorithms. First, the basic concepts of cultural algorithms will be presented. Then, more sophisticated extensions into the realms of multi-population approaches and dynamic optimization are surveyed.

A cultural algorithm as initially envisioned by \cite{reynolds1994introduction} extends the biologist model of a genetic algorithm with a linked concept of culture. Here, a so called belief system stores important information regarding the search. What precisely entails important information is problem dependent.  

For cultural algorithms to function, a communication protocol between the population component and the belief system is required. For once, this protocol must control write access to the belief system and also manages how the belief system influences the individuals. These functions are called update and influence functions, respectively. To extend the evolutionary notion of survival of the fittest, only the best individuals of any generation are permitted to update the belief system.

Herein lies also a potential pitfall: Since the best individuals are exerting a large influence over the entire population via the belief system, local optima can become deadly traps, especially in dynamic scenarios. In a problem where the data in the search space is updated frequently, the belief system must fear to stay behind and actually prevent its population from exploring more promising areas of the search space.

Cultural algorithms are a vital field of research. For instance, they have been daisy chained so that a second cultural algorithm optimizes the belief system obtained from an earlier run. This process, akin to two-pass encoding of media data was demonstrated by \cite{ostrowski2002using}. For an overview of genetic algorithms and their suitability for cultural algorithms, see \cite{becerra2004cultural}.

The extension of cultural algorithms to multi-population approaches is still in its early phase. For a general overview on multi-population evolutionary computing, see \cite{cantú1998survey,haupt1998practical,homayounfar2003advanced}. As a quasi-cultural algorithm with a rudimentary belief system, \cite{ray2003society} consider the employment of the mechanics of societal leadership for the purposes of a search. There, solutions cluster together and vote the best of their respective ranks to be a leader. These leaders further influence their constituents. The leaders themselves, after becoming leader are only influenced by other leaders. This approach is marked with a notion of cooperation between competing societies, where information is exchanged for a greater good.

\cite{digalakis2002multipopulation} employ a number of parallel populations, each with a separate belief system. The best solutions of each belief system are exchanged. \cite{coelho2009improved} extend that approach with having multiple belief systems control the mutation rates of their attached populations. In both of these solutions, there is some form of migration of individuals between belief systems, but the information exchange between belief systems themselves is restricted. \cite{guo2011novel} give an excellent account on their approach that only allows for belief system communication; migration of individuals is prohibited. Noteworthy here is, that they envision a complex protocol of merging the contents of different belief systems to form an optimal belief system content. This merged, or enriched content is then copied over to other belief systems.

\section{Evolving Evolution: Politics}
Culture is a basic treat of humanity. Among other things, culture includes norms of human behavior, limits of human behavior. By this, culture describes behavior that is acceptable and sets forth ways of punishing unacceptable behavior. The processes of negotiating acceptable behavior is also know as politics. If we can settle for politics to equate with that process, then the institutions where this process takes place should be named polity. And finally, the contents themselves are named policies in political science. With these terms, we can start conceptualizing culture as something extremely volatile, by elevating it to the levels of ideology. An ideology promises a greater good to its followers, as long as they adhere to the prescribed norms. By this definition, an ideology will have many followers as long as the encoded norms and values will provide the followers with an advantage over individuals subscribed to another ideology. 

Obviously, humankind has produced a great many number of cultures over the course of history. At a naive glance, all of them are equal. Yet, at times, some cultures are more persuasive than others. Let this dominant culture, or ideology be known as hegemon. A hegemonic culture seeks to eradicate opposition and instill its values, its content into every individual as universal. The methods of this instilling vary from one culture to the next, but the pattern appears to be constant. However, this only works as long as an ideology's promises are being kept.

When any ideology amasses enough power, i.e. enough followers to achieve the status of hegemon, that ideology provides usable solutions to real problems experienced by its followers by definition. Subjectively, the individuals judge the cost of subscribing to another ideology as being higher than to remain with the hegemonic ideology. Unfortunately, ideologies are exceptionally resistant to change. So since the world keeps on turning, and the answers an ideology provides by means of its encoded norms and values fail to provide its subscribers with a definite advantage, this ideology is waning. Its followers will start to look for alternatives to subscribe to, that provide better solutions for their needs. 

These dissidents form subcultures of their own, shielded from the hegemon by the use of code and clandestinity. There they develop alternative norms and values (through the means of politics). Eventually, once a dissident group becomes strong enough, they will abandon their hiding and challenge the hegemon. In the real world, this means revolution. Not always these revolutions are bound to be violent, and not always they are predetermined to be successful. But the message is clear: the hegemon has lost support in the population and its belief system is bound to be replaced by something new. 

For example, consider the revolutions reaping throughout Europe in and around the key year of 1848. The old ideology of absolute monarchical power had started to fail its followers. In clandestine circles at first, individuals were searching for alternative solutions that were more apt to the problems they faced. Eventually, the masses were taking to the streets, and at great loss of life, challenged the hegemon. What happened next, depended on the country. In some locations, the effective powers of the monarch were limited, in others the hegemon survived the altercation more or less intact.

There have always been multiple, concurrent cultures. These cultures can be conceived as ideologies, and at times, some of them may claim hegemonic and dominant status over others. Whenever a dominant ideology, a hegemon fails to deliver its followers from evil, so to speak, these followers will start to reorient themselves and follow alternative, dissident subcultures. Until, eventually, popular subcultures take over, and it all starts over again. Much of human development is found in this evolution of ideas. From the god-emperors of Rome, to absolute monarchs in medieval Europe to democracies and her challengers, ideas and revolutionary changes to them marked the path. In the next section, this human behavior is mapped onto an algorithm, ripe for applications in the optimization of dynamic problems. 

\section{Revolutionary Algorithms}
The proposed revolutionary algorithms seek to emulate the struggle of ideas for hegemonic position. A revolutionary algorithm is an extension to cultural algorithms with multiple sub-populations, each equipped with their own belief space. The novelty in this approach lies within the communication protocols between populations and belief spaces. To the best of our knowledge, we are the first to suggest this.

At the foundation of a revolutionary algorithm lies a genetic algorithm that steers the evolution of individuals. There is a large number of individuals that concurrently seek out the best solution. The genetics of their reproduction is guided by a belief system, to which a group of individuals adhere. Initially, the link between belief system and individual is random.

During the course of the optimization, individuals are more likely to subscribe to the belief system that is most successful. Once a belief system has lost all of its followers, it will be deleted. Since it is the belief system that an individual subscribes to has definite effect on the genetics of that individual's children, the effect of the belief system is directly visible in the population after a single generation.

So far, this approach is similar to a cultural algorithm with multiple sub-populations and individual migration. The distinct feature lies in the definition of success. In terms of genetic optimization, success is the distance between an individual's fitness and the true minimum. Since the true minimum is not known, a different metric is needed. We suggest to use the average rate of improvement over a certain time frame. So a population/belief system combination that achieves lower values at a greater rate than another one in a given time, the former will be considered more successful. Since the success of a belief system directly translates into the number of followers is has, the notion of choosing a suitable ideology is aptly covered.

A key notion of genetic algorithms is that in the beginning of the optimization process, improvements to the initially random solutions are provided rapidly. At later stages, improvements are only produced over the course of many generations. So in the proposed revolutionary algorithm, initially a hegemon -- dominant belief space -- will become evident. As the generations progress, the rate of improvement per generation of that belief system will wane, and individuals of the population will start to spawn alternative belief systems and start subscribing to them. 

The proposed algorithm starts out with multiple population and belief system combinations. These subcultures all compete for finding the best solution. Eventually, one culture will improve faster and produce the globally best individual and thus become the global hegemon. However, as this hegemonic subculture comes closer to the perfect solution, its rate of improvement will decrease. As the population being influenced by the hegemon's belief system, fails to witness further improvement, the probability of them being influenced by other subcultural belief systems increases. 

The longer the hegemon fails to produce significant improvement, the more subcultures will spawn, most of them with a higher rate of improvement (and thus attractiveness) than the hegemon. These dissident subcultures will lure population away from the hegemon until one of them takes over the role of hegemon. 

Since revolutionary algorithms are primarily intended to solve dynamic optimization problems, the number of dissident subcultures spawning is not only dependent on the lack of hegemonial improvement, but also from the detection of a change in data. The more data points change from one time to another, the more subcultures are spawned.

To operationalize human behavior as identified above, some form of communication protocol needs to be established. A way of determining the hegemon needs to be found, as well as a means of clarifying the allegiance of individuals to a belief system. The hegemon shall be the belief system/population combination that is largest, i.e. that has the most followers or whose population component is largest. The allegiance function is more complicated. It is a stochastic function that depends on the number of followers of any belief system, the rate of improvement of that belief system and gives the probability of an individual $j$ appertaining ($A$) to a belief system $i$: 
\begin{equation}
  P(A_{ij})=\frac{n_i r_i}{N r_j}
\end{equation}

with $n_i$ being the proportion of followers of that belief system out of all individuals, $r$ being the success of that belief system or that of the $j$th individual's belief system and $N$ the number of all belief systems in existence. The outcome of this function is the probability of an individual $j$ subscribing to a belief space $i$. This allegiance function is at the core of the proposed revolutionary algorithms. It contributes directly to the vitality of a belief system and the chance of revolution against a hegemon.

As the hegemon fails to produce a sufficient rate of improvement, the probability of occurrence of dissidents increases. These dissidents are elements of the hegemon's belief system's population. However, they forsake the hegemons influence and start to form their own, independent belief systems. The spawning of dissidents is a side product of evaluating the allegiance function described above: The smaller the attractiveness of the hegemon, the larger the probability of dissidents spawning.

\begin{equation}
  P(S)=dp \times div  %\frac{max(d_ij)}{\sum{d_ij}}}
\end{equation}

This formula gives the probability of dissident sub-populations spawning. It is based on some measure $dp$ quantifying the severeness of a data change in the search space, for instance the proportion of data points that have changed during a certain time frame. The factor $div$ is some measure of diversity: The more diverse the different sub-populations are, the more active they are in different regions of the search space, the less probable is the spawning of new dissidents. A possible way of arriving at this quantity is $$div=\frac{\max(d_{mn})}{0.5\,\sum_{m\neq n} d_{mn}},$$ with $d_{mn}$ being the entries of a euclidean distance matrix between any two solutions $m$ and $n$. 

\begin{algorithm}[htb]
\caption{Pseudocode of a revolutionary algorithm.}  \label{alg:Hochreiter-RAcode}
\begin{algorithmic}[1]
\STATE Initialize() $r$ random populations and these individuals' allegiences
\STATE Estimate fitness of $i$ individuals in $r$ populations
\STATE Update $b$ belief systems according to Update() and Allegience() functions
\WHILE{not stopCriterion}
   \STATE Apply Influence() function with respect to belief system and individual
   \STATE Procreate offspring
   \STATE Evaluate fitness of offspring
   \STATE Update belief systems according to Update() and Allegience() functions
   \STATE Spawn new dissidents
   \STATE Update individuals allegience
\ENDWHILE
\STATE Report best individual
\end{algorithmic}
\end{algorithm}

The proposed algorithm's pseudocode is presented in Fig. \ref{alg:Hochreiter-RAcode}. In words, as a first step the $r$ different populations and their -- initially random -- allegiances are initialized. At the beginning, the belief spaces are still empty. As a next step, every individual $i$'s fitness is being determined and their respective belief spaces are being updated with that knowledge. After the initialization, the optimization starts, until some stop criterion (in dynamic optimization problems, usually the end of data input), is met.

In the optimization phase, first, the cultural algorithm's influence function is applied to directly alter the individuals' behavior with respect to their belief space. Then offspring is procreate as informed by the belief space and this offspring's fitness evaluated. The best individuals are of any generation are then able to update the belief system according to their respective allegiance functions and the global update function. After the belief system has been updated, the allegiance system is re-evaluated if there are any shifts to be recorded. According to the hegemon's performance the availability of new data points, dissidents are spawned. Finally, an individual's allegiance to a given belief system is being updated. This optimization process is continued until some stop criterion is met. Typically, in a dynamic optimization problem, this criterion would be the end of data input.

\begin{figure}
  \centering
  \includegraphics[width=0.75\textwidth]{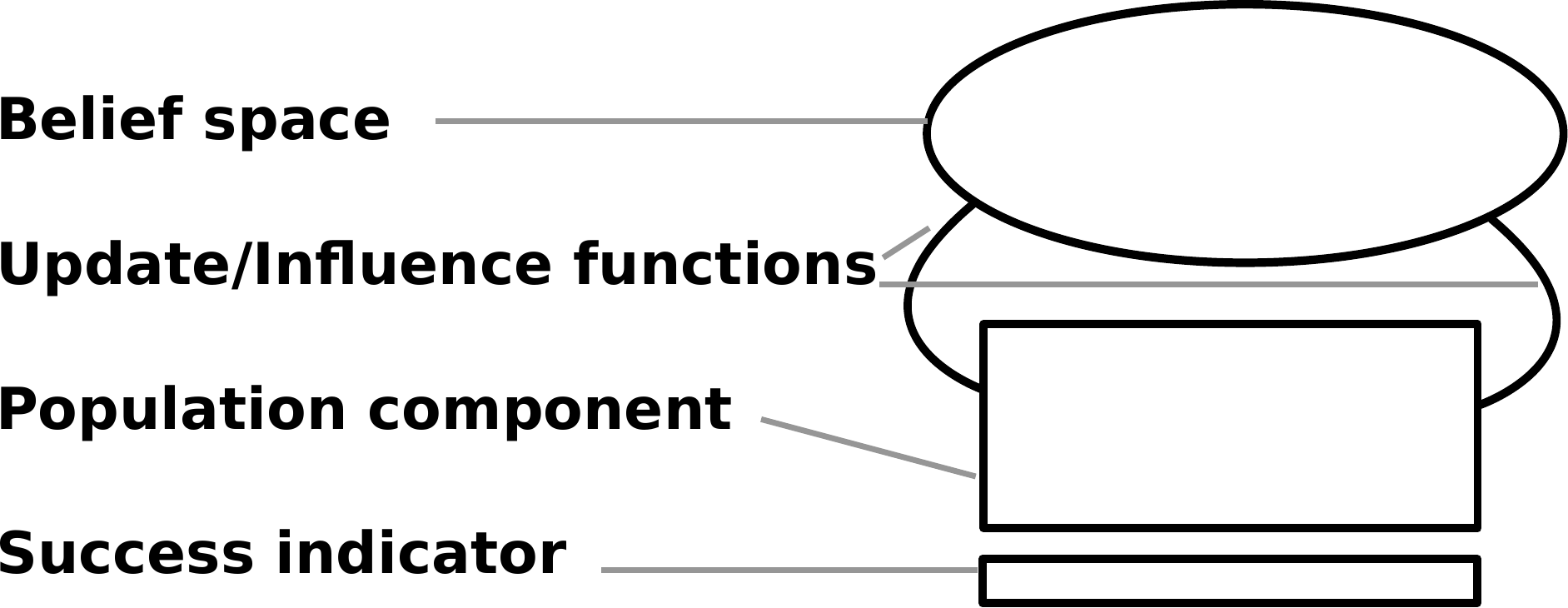}
  \caption{\label{fig:Hochreiter-pict} Building block of revolutionary algorithms: the components of a population/belief system combination.}
\end{figure}

\begin{figure}
  \centering
  \includegraphics[width=0.75\textwidth]{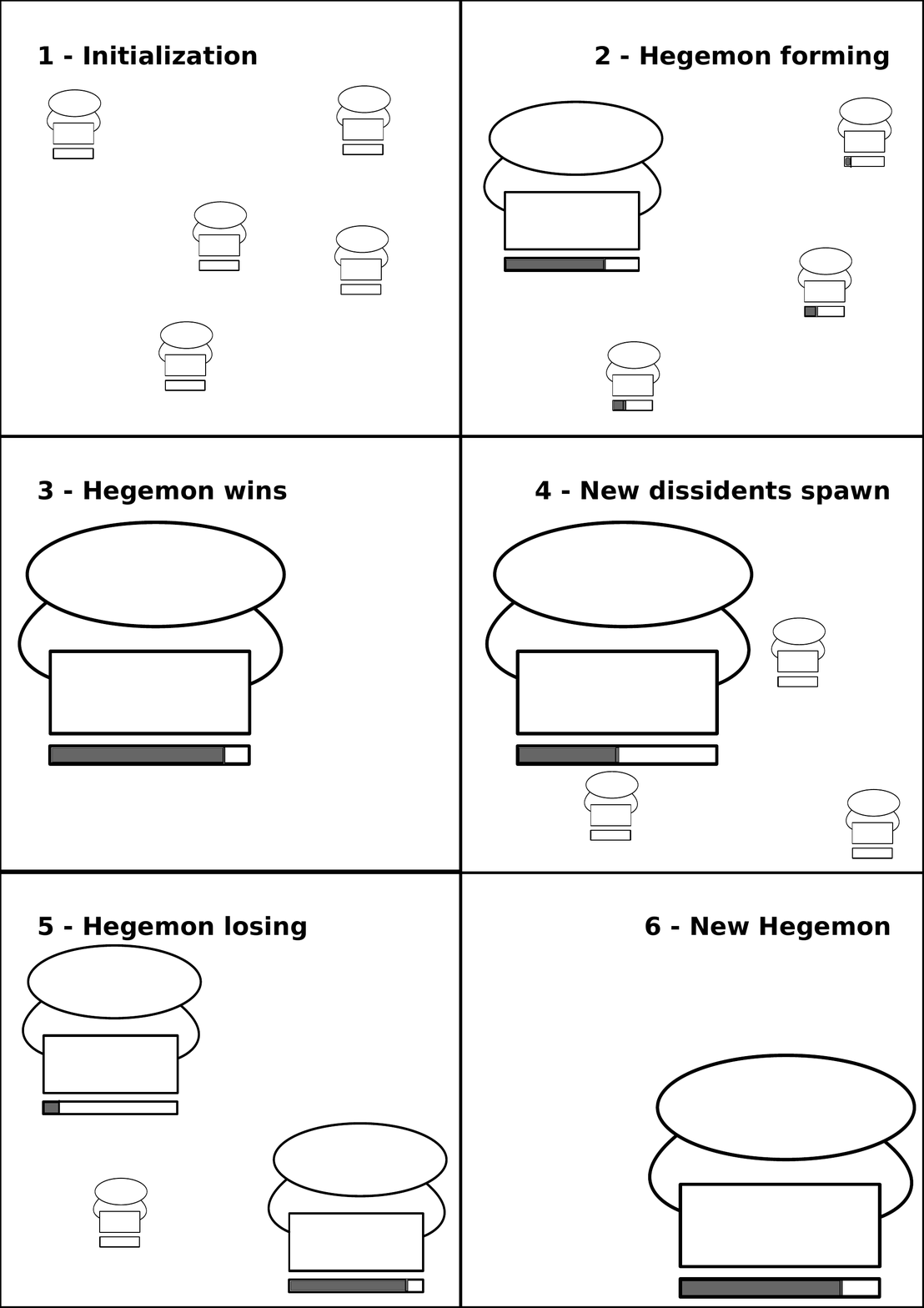}
  \caption{\label{fig:Hochreiter-proc} Six typical stages during an optimization run using a revolutionary algorithm.}
\end{figure}

Fig. \ref{fig:Hochreiter-proc} depicts the process of how a revolutionary algorithm works. The six panels depict six typical stages during an optimization run. The pictogram used is explained in Fig. \ref{fig:Hochreiter-pict}. At first different sub-populations are initialized together with their belief systems. Here, all the population/belief system combinations have the same size. After a few generations, the forming of a hegemon begins: a single combination of population and belief system will outclass the others, as depicted with the success indicator. By being more successful than the competing subcultures, population is drawn to this hegemon. In panel 3, all population has subscribed to the most successful belief system. Now the algorithm either stops, as some stopping criterion is met, or the optimization continues, potentially with new data being introduced. As the hegemon will fail to produce sufficient increase in fitness to keep its followers at bay, dissident sub-cultures will form. Eventually, 
one of this sub-cultures will become popular among the population and draw individuals away from the (former) hegemon, and become a hegemon itself (panels 5 and 6).

\section{Conclusion}
The proposed evolutionary algorithm, Revolutionary algorithm, seeks to emulate human political behavior. Conceptionally, it is thus an extension of cultural algorithms mapping the competition of different human cultures onto the heuristical determination of an optimum. By doing so, the threat of any cooperation -- to peacefully reside at a place, just because a better one is not dared to be imagined, is avoided. Thus, the search can quickly and confidently adopt to updated search spaces and is less prone to get trapped in local optima. As opposed to a simple restart of the search, a revolutionary algorithm keeps track of the previous search history. As suggestions for future work we would like to point to researching runtime behavior of the algorithm in different applications and on combining the idea of multiple, competing belief systems with different genetic algorithms for the population space. 

\bibliography{ra}
\bibliographystyle{plain}

\end{document}